\documentclass[conference]{IEEEtran}
\IEEEoverridecommandlockouts
\usepackage{booktabs} 
\usepackage{graphicx}
\usepackage{setspace}
\usepackage{amsmath}
\usepackage{dsfont}
\usepackage{amsfonts}
\usepackage{lipsum}
\usepackage{multicol}
\usepackage{float}
\usepackage{bm}
\usepackage{color}
\usepackage[dvipsnames]{xcolor}
\usepackage{etoolbox}
\usepackage{soul}
\usepackage{amssymb}
\usepackage[linesnumbered,ruled,vlined]{algorithm2e}
\usepackage{mathtools,amssymb}
\usepackage{caption}
\usepackage{float}
\usepackage{amsthm}
\usepackage[caption = false,subrefformat=parens,labelformat=parens]{subfig}
\usepackage{adjustbox}
\usepackage{afterpage}
\usepackage{multirow}
\usepackage{mathrsfs}

\def\BibTeX{{\rm B\kern-.05em{\sc i\kern-.025em b}\kern-.08em
    T\kern-.1667em\lower.7ex\hbox{E}\kern-.125emX}}
\begin{document}

\title{Generative AI-Powered Plugin for Robust\\ Federated Learning in Heterogeneous IoT Networks

\author{
Youngjoon Lee\textsuperscript{\rm 1}, Jinu Gong\textsuperscript{\rm 2}, Joonhyuk Kang\textsuperscript{\rm 1} \\
\textsuperscript{\rm 1}School of Electrical Engineering, KAIST, South Korea\\
\textsuperscript{\rm 2}Department of Applied AI, Hansung University, South Korea\\
Email: yjlee22@kaist.ac.kr, jinugong@hansung.kr, jkang@kaist.ac.kr
}

\thanks{This research was supported by the Institute of Information \& Communications Technology Planning \& Evaluation (IITP)-ITRC (Information Technology Research Center) grant funded by the Korea government (MSIT)  (No.RS-2025-02309685, Development of Programmable
Infrastructure Technology for Guaranteed Application Performance).\\
}
}

\maketitle

\begin{abstract}
Federated learning enables edge devices to collaboratively train a global model while maintaining data privacy by keeping data localized. 
However, the Non-IID nature of data distribution across devices often hinders model convergence and reduces performance.
In this paper, we propose a novel plugin for federated optimization methods that approximates Non-IID data distributions to IID through generative AI-enhanced data augmentation and balanced sampling strategy. The key idea is to synthesize additional data for underrepresented classes on each edge device, leveraging generative AI to create a more balanced dataset across the FL network. Additionally, a balanced sampling approach at the central server selectively includes only the most IID-like devices, accelerating convergence while maximizing the global model's performance. Experimental results validate that our approach significantly improves convergence speed and robustness against data imbalance, establishing a flexible, privacy-preserving FL plugin that is applicable even in data-scarce environments.
\end{abstract}

\section{Introduction}
\label{sec:intro}
Deep learning \cite{goodfellow2016deep} has shown remarkable performance in various domains by learning complex representations from large datasets.
However, achieving such performance often requires extensive data, which is impractical in many real-world scenarios due to privacy concerns and the prohibitive costs associated with data sharing \cite{simeone2022machine}. 
Federated learning (FL) \cite{mcmahan2017communication} addresses these challenges by enabling edge devices to train models locally and share only the updated models with a central server, thus maintaining data privacy and security \cite{khan2021federated, lee2024security}. 
Therefore, FL provides a crucial solution for privacy-preserving sector, especially in healthcare \cite{lee2025revisit}.

While FL achieves comparable performance to centralized learning, it often requires numerous global epochs to achieve satisfactory accuracy \cite{zhou2018convergence, lee2023fast}.
This challenge intensifies when data at each edge device is non-independent and identically distributed (Non-IID) \cite{zhao2018federated} or when computational capabilities vary across devices \cite{li2020federated}. 
Such statistical and computational heterogeneity among edge devices often results in local models drifting towards distinct minima, thus degrading the performance of the aggregated model \cite{kairouz2021advances} and increasing the communication overhead required for convergence \cite{ding2022federated}, \cite{lee2022accelerated}. 
Addressing these issues is essential for deploying FL effectively in real-world applications \cite{li2019convergence}, particularly in medical tasks where data distributions are highly imbalanced \cite{li2021comprehensive}.

\begin{figure}
     \centering
     \includegraphics[width=\columnwidth]{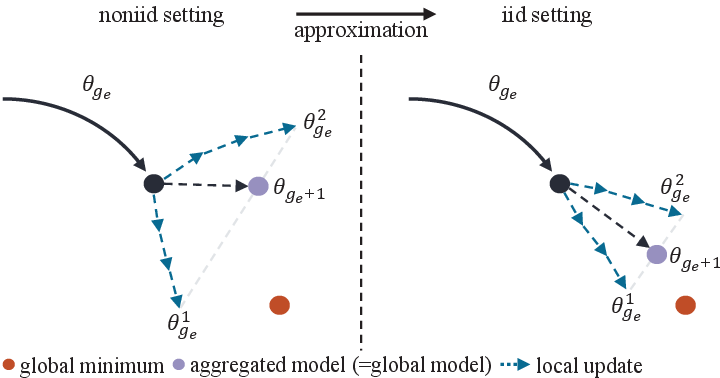}
     \caption{Illustration of FL in a heterogeneous environment with the proposed plugin: This approach approximates Non-IID data distributions to IID, balancing statistical heterogeneity across edge devices to enhance convergence.}
     \label{fig:pic0}
\end{figure}

Several methods have been proposed to address this client-drift phenomenon by modifying the local training process. 
For instance, proximal regularization \cite{li2018federated} and restricted-softmax function \cite{li2021edrs} help alleviate client-drift but increase edge-side complexity.
In this paper, we propose a novel plugin to mitigate client drift in heterogeneous FL environments.
Our approach involves paraphrasing data using generative AI and selecting an optimal set of edge devices to participate in the updates, thereby approximating the Non-IID data distributions to IID as shown in Fig. \ref{fig:pic0}.
Moreover, the proposed plugin can be seamlessly integrated with any FL method.

The main contributions of this paper are as follows:
\begin{itemize}
    \item We propose a generative AI method called \emph{Non-IID to IID Approximation} for FL.
    \item We enable edge devices to synthesize data for underrepresented classes to balance the dataset.
    \item We design a central server strategy to select the most balanced devices for aggregation.
    \item We numerically show that our plugin improve training speed and accuracy in medical text dataset.
\end{itemize}

\section{Preliminaries}\label{sec:preliminaries}
In this section, we summarize the commonly used FL methods.
We denote $N$ as the number of participating edge devices and $g_e$ as the global epoch for aggregation.

\textbf{FedAvg:}  
In FedAvg, each device trains its model on local data and sends the updated parameters to a central server, which averages them to form a new global model.
\bigskip

\textbf{FedProx:}  
FedProx adds a proximal term to the local training objective to keep updates close to the global model  $\theta$ and reduce client drift.
Each device $n$ minimizes its local objective function $f_n(\theta)$ as:
\[
f_n(\theta_n) = f_n(\theta_n) + \frac{\mu}{2} \lVert \theta_n - \theta \rVert^2,
\]
where $\mu$ controls the proximal strength.

\bigskip

\textbf{FedRS:}  
FedRS scales down updates for classes missing in a device's data. The adjusted probability for class \( y \) is:
\[
\pi_{i,y} = \frac{\exp(\alpha_y\, w_y^T h_i)}{\sum_{j=1}^{\mathcal{C}} \exp(\alpha_j\, w_j^T h_i)},
\]
with \( \alpha_y = 1 \) if class \( y \) is present and \( \alpha_y = \alpha \in [0,1] \) if absent. 
This reduces the influence of missing classes, aiding convergence in Non-IID environments.

\section{Problem and Model}
\label{sec:main}
\subsection{Federated Heterogeneous Setting} 
We consider a federated network composed of $N$ edge devices, each communicating with a central server. 
Each edge device $n=1, \ldots, N$ holds a private dataset $\mathcal{D}^{n}$, with each dataset varying in both size and class distribution.
Furthermore, edge devices possess different computational capabilities, affecting the number of local epochs $1\leq l_e \leq L$ within maximum value $L$, they can perform during each update round. 
FL aims to optimize a shared global model $\theta_{g_e} \in \Theta$ within the parameter space $\Theta$ without transferring the local datasets $\mathcal{D}=\cup_{n=1}^{N}\mathcal{D}^{n}$ to the central server
Thus, the data privacy of each device is preserved.

To achieve this, the FL objective aims to minimize the global loss function $F(\theta)$, defined as:
\begin{equation}
\min_{\theta_{g_e} \in \Theta} F(\theta) \triangleq \frac{1}{N} \sum_{n=1}^{N} f_{n}(\theta_{g_e}^n),
\end{equation}
where $F(\theta_{g_e})$ represents the global objective at the central server, and each local objective function $f_n(\theta)$ for edge device $n$ is defined as:
\begin{equation}
f_{n}(\theta_{g_e}^n) = \frac{1}{|\mathcal{D}^{n}|} \sum_{(x, y) \in \mathcal{D}^{n}} f_{n}(\theta_{g_e}; x; y).
\end{equation}

Here, $f_{n}(\theta; x)$ represents the loss function for data sample $x$ within the local dataset $\mathcal{D}^{n}$.
Due to the Non-IID nature, the distribution of data on edge devices $i$ and $j$ may differ significantly for $i \neq j$.

\subsection{Federated learning with Proposed Plugin}
\begin{figure}[t]
    \centering
    \includegraphics[width=\columnwidth]{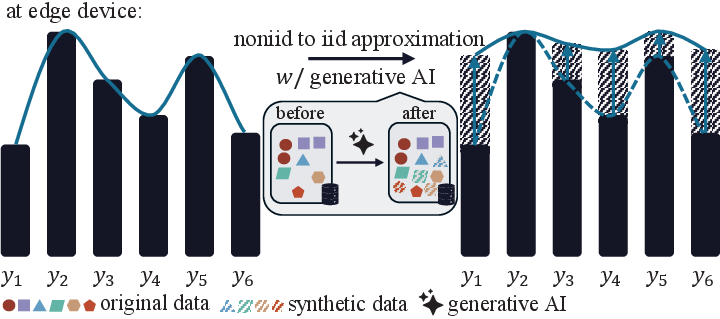}
    \caption{Illustration of data augmentation at edge device using generative AI to approximate Non-IID data to IID distributions. Original data is supplemented with synthetic data generated by generative AI to balance class distributions.}
    \label{fig:pic2}
\end{figure}

\subsubsection{Phase 1-Data Augmentation with Generative AI}
As shown in Fig. \ref{fig:pic2}, we utilize generative AI to alleviate data imbalance on edge devices.
By generating synthetic data, especially for underrepresented classes, the technique augments the datasets, enabling a transformation from Non-IID distributions to approximate IID conditions. Providing a practical solution for edge devices with limited or skewed datasets, this data synthesis method helps to achieve a more balanced learning environment.

Let $\mathcal{D}^n = \{(x_i^n, y_i^n)\}_{i=1}^{|\mathcal{D}^n|}$ represent the local dataset at edge device $n$, where $x_i^n$ is the input data, and $y_i^n$ is the corresponding label.
When certain classes are underrepresented in $\mathcal{D}^n$, a generative AI $\mathcal{G}$ is used to synthesize additional samples to balance the dataset:
\begin{equation}
\mathcal{D}^n_{\text{aug}} = \mathcal{D}^n \cup \mathcal{G}(\mathcal{D}^n),
\end{equation}
where $\mathcal{G}(\mathcal{D}^n)$ generates synthetic samples to balance the distribution. Specifically, the generative AI model produces samples for each class \( y \) where data is insufficient, ensuring that class imbalance is minimized. To avoid introducing unseen data, the process only generates data for classes that are present in $\mathcal{D}^n$:
\begin{equation}
\mathcal{G}(\mathcal{D}^n) = \bigcup_{y \in \mathcal{C}, |\mathcal{D}^n_y| > 0} \mathcal{G}_y(\mathcal{D}^n),
\end{equation}
where $\mathcal{G}_y(\mathcal{D}^n)$ denotes the synthetic data generated for class \( y \), and \( \mathcal{C} \) represents all classes.

To determine the quantity of synthetic data, we calculate the deficiency ratio for each class \( y \) by comparing its size \( |\mathcal{D}^n_y| \) with the maximum class size \( |\mathcal{D}^n_{\text{max}}| \):
\begin{equation}
\Delta_y = \frac{|\mathcal{D}^n_{\text{max}}| - |\mathcal{D}^n_y|}{|\mathcal{D}^n|}, \quad \text{if } |\mathcal{D}^n_y| > 0.
\end{equation}

If \( \Delta_y \) is positive, indicating class imbalance, the generative model synthesizes additional samples as:
\begin{equation}
\mathcal{G}_y(\mathcal{D}^n) = \begin{cases} 
r(1, \Delta_y \cdot |\mathcal{D}^n|) & \text{if } |\mathcal{D}^n_y| > 0 \\
0 & \text{if } |\mathcal{D}^n_y| = 0
\end{cases},
\end{equation}
where \( r(1, \Delta_y \cdot |\mathcal{D}^n|) \) is a random integer between 1 and $\Delta_y \cdot |\mathcal{D}^n|$. The final augmented dataset for each edge device $n$ is:
\begin{equation}
\mathcal{D}^n_{\text{aug}} = \mathcal{D}^n \cup \bigcup_{y \in \mathcal{C}} \mathcal{G}_y(\mathcal{D}^n).
\end{equation}

Data augmentation ensures that the local model trained on $\mathcal{D}^n_{\text{aug}}$ benefits from a more balanced class distribution, improving the global model’s performance by mitigating the effects of Non-IID nature.

\begin{figure}[t]
    \centering
    \includegraphics[width=\columnwidth]{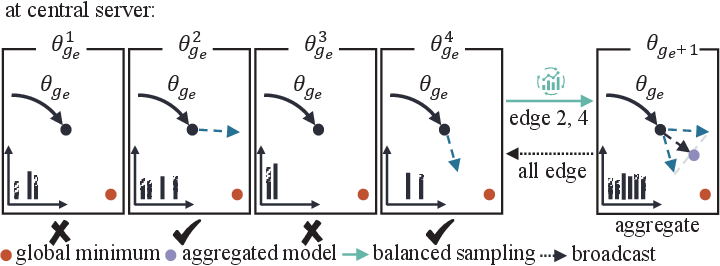}
    \caption{Illustration of the proposed balanced sampling at the central server, selecting set of $K=2$ edge devices with statistically representative data distributions for aggregation to achieve a optimized global model.}
    \label{fig:pic3}
\end{figure}

\subsubsection{Phase 2-Balanced Sampling}
Following data augmentation phase, the central server implements a balanced sampling strategy as shown in Fig \ref{fig:pic3}.
This strategy optimizes model aggregation at the central server by focusing on edge devices whose set of data distributions most closely approximate an IID target.
For each edge device $n$ with augmented dataset $\mathcal{D}^n_{\text{aug}}$, the proportion of data for each class $y$ is calculated as:
\begin{equation}
p_n(y) = \frac{|\{x \in \mathcal{D}^n_{\text{aug}} : y = y\}|}{|\mathcal{D}^n_{\text{aug}}|},
\end{equation}
where $p_n(y)$ denotes the proportion of class $y$ in edge device $n$'s dataset. By collecting these class proportions from each device, the central server calculates a weighted global distribution $p_{\text{global}}(y)$ as:
\begin{equation}
p_{\text{global}}(y) = \frac{\sum_i p_i(y) \cdot |\mathcal{D}^i_{\text{aug}}|}{\sum_i |\mathcal{D}^i_{\text{aug}}|}.
\end{equation}
Then, the central server measure the distance between each edge device's class distribution $p_i(y)$ and $p_{\text{global}}(y)$ using a distance function as:
\begin{equation}
d(p_i(y), p_{\text{global}}(y)) = \| p_i(y) - p_{\text{global}}(y) \|_2.
\end{equation}

Next, the central server then forms a subset by selecting $K$ edge devices that show the smallest distances to $p_{\text{global}}(y)$. 
Only these selected edge devices participate in the local learning stage, ensuring that the aggregation process draws from a well-balanced subset of data distributions. 

After local training, the central server aggregates the locally updated models $\vartheta = \{\theta^{k}_{g_e}\}_{k=1}^{K}$to update the next global model $\theta_{g_e+1}$ as follows:
\begin{equation}
\theta_{g_e+1} = \frac{\eta_g}{K}\sum_{k=1}^{K}\theta_{g_e}^{k}, 
\end{equation} where $\eta_g$ is the global learning rate.
This process is repeated for each global epoch $g_e$, until the desired performance is achieved or the maximum number of global epochs $G$ is reached. 
Overall procedure is described in Algorithm \ref{alg_1}.

\begin{algorithm}
\DontPrintSemicolon
\SetKwFunction{GenerateSyntheticData}{GenerateSyntheticData}
\SetKwFunction{ComputeDeficiencyRatio}{ComputeDeficiencyRatio}
\SetKwFunction{ComputeClassProportions}{ComputeClassProportions}
\SetKwFunction{ComputeDistanceToIID}{ComputeDistanceToIID}
\SetKwFunction{SelectDevices}{SelectDevices}
\SetKwFunction{AggregateModels}{AggregateModels}

\textbf{Initialize:} Global model $\theta_{0}$, maximum epochs $G$\;

\For{global epoch $g_e = 0$ \KwTo $G-1$}{
    \tcc{Phase 1: Data Augmentation}
    \For{each edge device $n = 1$ \KwTo $N$ \textbf{in parallel}}{
        \For{each class $y$ where $|\mathcal{D}^n_y| > 0$}{
            Compute deficiency ratio $\Delta_y$\;
            \If{$\Delta_y > 0$}{
                Generate synthetic data $\mathcal{G}_y(\mathcal{D}^n)$\;
            }
        }
        Form augmented dataset $\mathcal{D}^n_{\text{aug}}$\;
    }
    \tcc{Phase 2: Balanced Sampling}
    \For{each device $n = 1$ \KwTo $N$}{
        Compute weighted global distribution  $p_{\text{global}}(y)$\;
        Compute class proportions $p_n(y)$\;
        Compute distance $d_n = \| p_n(y) - p_{\text{global}}(y) \|_2$\;
    }
    Select $K$ edge devices with smallest $d_n$\;
    \textbf{Broadcast:} Send $\theta_{g_e}$ to selected $K$ devices\;
    \tcc{Local Training}
    \For{each device $s_k \in \mathcal{S}$ \textbf{in parallel}}{
        Initialize $\theta_{g_e}$ to local model\;
        \For{local epoch $l_e = 1$ \KwTo $L$}{
            $\theta_{g_e, l_e+1} = \theta_{g_e, l_e} - \eta_{l} \nabla f_{k}(\theta^{k}_{g_e, l_e})$\;
        }
    }
    \tcc{Aggregation at Central Server}
    Receive models $\{\theta^{k}_{g_e}\}_{k=1}^K$\;
    Update $\theta_{g_e+1} = \frac{\eta_g}{K} \sum_{k=1}^{K} \theta^{k}_{g_e}$\;
}
\textbf{Output:} Final global model $\theta_{G}$\;

\caption{FL with Proposed Plugin}
\label{alg_1}
\end{algorithm}
\vspace{-0.4cm}

\section{Experiment and Results}\label{sec:experiment}
\subsection{Experiment Setting}
We validate our plugin by applying it to common FL algorithms on a medical text classification task \cite{schopf2022evaluating} using Intel’s Gaudi 2 AI accelerator.
The text classification models were trained on a Non-IID distribution across $N = 100$ edge devices \cite{hsu2019measuring}, with $K = 10$ devices selected in each global round.
Additionally, We used the Ollama API for data augmentation by leveraging multiple representative LLMs to generate synthetic data from the original input, including Gemma 1.0 (7B) \cite{team2024gemma1}, Gemma 2.0 (2B and 9B) \cite{team2024gemma2}, Llama 3.1 (8B), Llama 3.2 (1B and 3B) \cite{dubey2024llama}, Phi 2 (3B) \cite{javaheripi2023phi}, Phi 3 (3B), and Phi 3.5 (3B) \cite{abdin2024phi}.
Further details on hyperparameter settings and experimental configurations are available in the open-source repository\footnote{https://github.com/NAVER-INTEL-Co-Lab/gaudi-fl}.

\subsection{Results}
\subsubsection{Impact on AI Model Diversity}
\begin{figure}[t]
  \centering
  \includegraphics[width=\columnwidth]{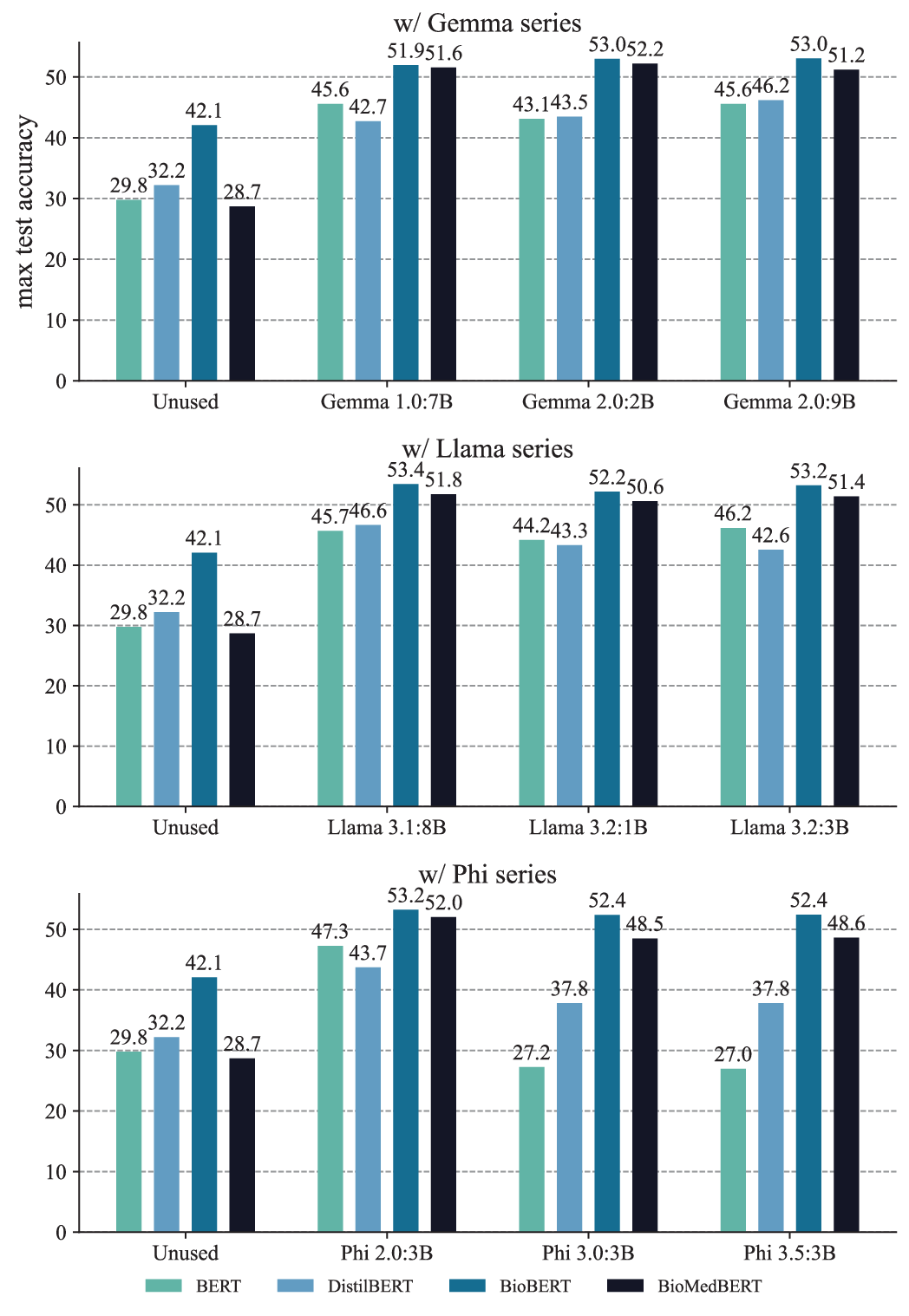}
  \caption{Performance comparison of text classification models with and without our proposed plugin.}
  \label{fig:experiment1}
\end{figure}

In this experiment, we evaluated our plugin's ability to enhance text classification models in FL environments. 
Moreover, we compared four models: general-purpose BERT \cite{kenton2019bert} and DistilBERT \cite{sanh2019distilbert}, and domain-specific BioBERT \cite{lee2020biobert} and BioMedBERT \cite{chakraborty2020biomedbert}. 
As shown in Fig. \ref{fig:experiment1}, performance evaluation showed significant improvements, with BERT's accuracy increasing from 29.8\% to 47.3\% and DistilBERT's from 32.2\% to 46.6\%. In addition, the medical domain pre-trained models achieved even greater gains, with BioBERT rising from 42.1\% to 53.4\% and BioMedBERT from 28.7\% to 52.0\%. 
Thus, based on these results, we use BioBERT as the edge device model for the subsequent experiment.

\subsubsection{Impact on Various FL Methods}
\begin{figure}[t]
   \centering
   \includegraphics[width=\columnwidth]{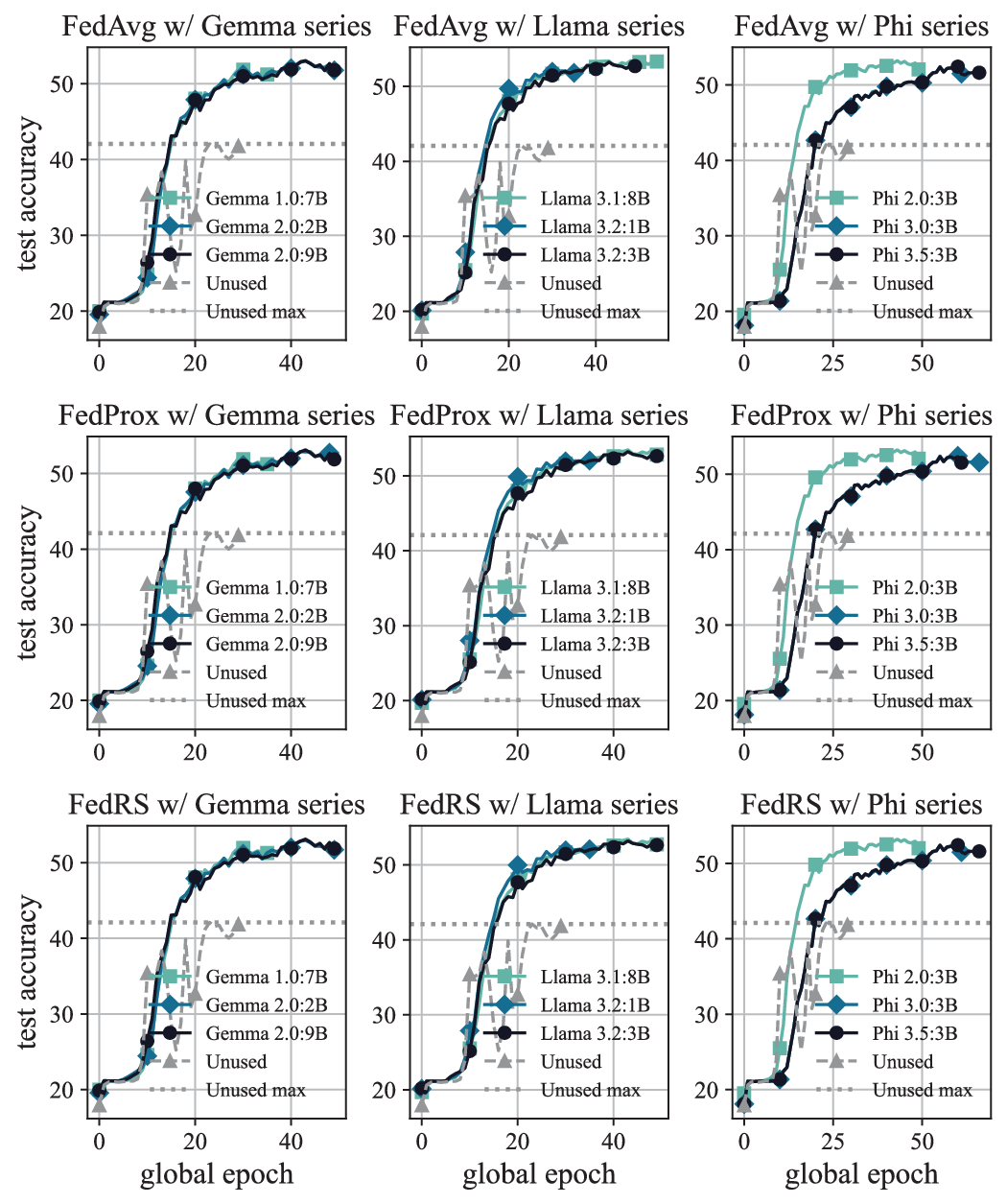}
   \caption{Performance comparison of different FL algorithms enhanced with the proposed plugin.}
   \label{fig:experiment2}
\end{figure}
To demonstrate the versatility of our proposed plugin, we examined its impact on convergence speed and overall performance across various FL optimization methods. 
As shown in Fig. \ref{fig:experiment2}, the FL methods enhanced with our plugin reached stable performance significantly faster than their vanilla counterparts. 
In fact, while the vanilla FL methods required 23 epochs to attain maximum accuracy, configurations with our plugin consistently converged between epochs 15 and 20. 
Notably, models such as Gemma 2.0:9B, the Llama 3.2 series, and Phi 2.0:3B achieved convergence as early as epoch 15, reflecting an approximate 35\% reduction in training epochs. 
Therefore, the results indicate that the proposed synthetic data augmentation and balanced sampling strategy improves both the convergence efficiency and global accuracy in FL.

\section{Conclusion}\label{sec:conclusion}
In this paper, we proposed a novel plugin for FL methods that addresses data heterogeneity challenges through generative AI-enhanced data augmentation and balanced sampling strategy.
In particular, the plugin balances local data distributions through generative augmentation and selects statistically representative devices for aggregation, improving robustness.
The plugin showed strong performance improvements in medical domain-specialized models, with BioBERT achieving 53.4\% accuracy from a 42.1\% baseline. 
Moreover, we numerically demonstrated the plugin's effectiveness across FL methods while reducing required training epochs. 
These results establish our plugin as a flexible solution for enhancing FL performance in real-world heterogeneous environments.
As future work, we plan to extend this plugin to multimodal FL settings to further broaden its applicability.

\bibliographystyle{IEEEtran}
\bibliography{reference}

\end{document}